\documentclass[letterpaper, 10 pt, conference]{ieeeconf}  

\IEEEoverridecommandlockouts                              

\usepackage{graphics} 
\usepackage{epsfig} 
\usepackage{amsmath} 
\usepackage{amssymb}  
\usepackage[table]{xcolor}

\title{\LARGE \bf
A Novel Robot Hand with Hoeckens Linkages and Soft Phalanges for Scooping and Self-Adaptive Grasping in Environmental Constraints*
}

\author{Wentao Guo, Yizhou Wang and Wenzeng Zhang, \textit{Member, IEEE}
\thanks{*Research supported by Foundation of \textit{Enhanced Student Research Training (E-SRT)} and Foundation of \textit{Open Research for Innovation Challenges (ORIC)}, X-Institute.}
\thanks{Wentao Guo is with Computer Science and Technology, Beijing Institute of Technology, China and Laboratory of Robotics, X-Institute, Shenzhen, China(email:yinsumirage@gmail.com).}
\thanks{Yizhou Wang is with the Southern University of Science and Technology, Shenzhen, China and Laboratory of Robotics, X-Institute, Shenzhen, China.}
\thanks{Wenzeng Zhang is with Laboratory of Robotics, X-Institute, Shenzhen, China and Dept. of Mechanical Engineering, Tsinghua University, Beijing, China (Corresponding author, email: wenzeng75@163.com).}
}

\begin{document}

\maketitle
\thispagestyle{empty}
\pagestyle{empty}

\begin{abstract}

This paper presents a novel underactuated adaptive robotic hand, Hockens-A Hand, which integrates the Hoeckens mechanism, a double-parallelogram linkage, and a specialized four-bar linkage to achieve three adaptive grasping modes: parallel pinching, asymmetric scooping, and enveloping grasping. Hockens-A Hand requires only a single linear actuator, leveraging passive mechanical intelligence to ensure adaptability and compliance in unstructured environments. Specifically, the vertical motion of the Hoeckens mechanism introduces compliance, the double-parallelogram linkage ensures line contact at the fingertip, and the four-bar amplification system enables natural transitions between different grasping modes. Additionally, the inclusion of a mesh-textured silicone phalanx further enhances the ability to envelop objects of various shapes and sizes. This study employs detailed kinematic analysis to optimize the push angle and design the linkage lengths for optimal performance. Simulations validated the design by analyzing the fingertip motion and ensuring smooth transitions between grasping modes. Furthermore, the grasping force was analyzed using power equations to enhance the understanding of the system's performance.Experimental validation using a 3D-printed prototype demonstrates the three grasping modes of the hand in various scenarios under environmental constraints, verifying its grasping stability and broad applicability.

\end{abstract}

\section{INTRODUCTION}

Over the past few decades, robotic end-effectors have undergone remarkable development, evolving from rigid industrial clamps to bionic systems capable of adaptive manipulation. This evolution is aimed at meeting the needs of operating in unstructured environments, where traditional fully-actuated mechanisms perform poorly. 

Dexterous hands, represented by the Utah/MIT Hand's pneumatic tendon system \cite{MIThand} and the DLR Hand's modular finger design \cite{DLRhand2008}, achieve remarkable manipulation versatility through independent joint control. The Stanford/JPL Hand's 9 degrees of freedom (DOF) \cite{JPLhand1987} enabled precision tasks, while NASA's Robonaut Hand \cite{Robonauthand} demonstrated tool manipulation in space environments. However, the intrinsic limitations of these systems, such as high complexity, excessive weight, and high cost, have spurred the emergence of underactuated mechanisms that sacrifice active DOFs for mechanical intelligence.

In the early stage, underactuated hands were divided into three modes: coupled, parallel, and self-adaptive. Subsequently, the coupled self-adaptive mode (COSA mode) and the parallel self-adaptive mode (PASA mode) were further proposed\cite{PASA2016}. Some well-known representatives include: the SARAH hand, the first hand used on the Space Station \cite{SARAH}, the Barrett hand that achieves adaptability with four motors \cite{Barrett}, the Robotiq hand where two fingers share one motor \cite{robotiq}, which realizes PASA mode and the SDM hand, an underactuated robot hand with four fingers \cite{SDM}. 

The PASA mode addresses the shortcoming of traditional parallel grasping robotic hands. When these hands perform grasping, the end-effector trajectory is circular, making it impossible to grasp thin-plate objects. By introducing various parallel mechanisms, parallel movement of the end-effector can be achieved, such as the Chebyshev linkage \cite{chebyshev2017}, the Hoeckens Mechanism \cite{Liu2019}, the Differential and Watt Linkages \cite{watt}, and Rack-Crank-Slider Mechanism\cite{Feng2024}. Yoon used Hart linkage mechanism coupled with a parallelogram to realize a fully passive robotic finger adaptable to the environment \cite{Kim2022}.

\begin {figure}[h]
    \centering
    \includegraphics [width=0.83\linewidth]{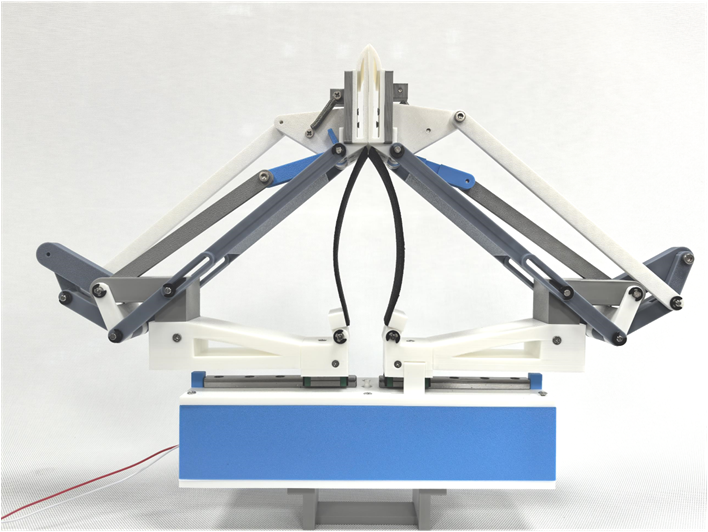}
    \caption {Hockens-A Hand.}
    \label {fig:show}
\end {figure}

This paper presents the Hockens-A Hand, a novel underactuated adaptive robotic hand equipped with a dual-parallelogram Hoeckens mechanism, which addresses various manipulation challenges through geometric innovation.

The key contributions of the Hockens-A Hand are as follows. Its co-axial linkage configuration, controlled by a single actuator, enables three grasping modes. For thin objects, it uses asymmetric scooping; for thick objects, symmetric scooping and enveloping grasps are employed, while direct parallel pinching is used for normal-sized items. Moreover, the hand can passively shift between these modes due to the intrinsic constraints of the linkage, allowing adaptive grasping under environmental constraints, such as contact with a table or irregular surfaces.To validate its practical performance, a 3D-printed prototype was constructed, demonstrating excellent grasping capabilities and robust adaptability to environmental constraints.

\section{DESIGN}

Parallel robot hands employing vertical fingertip paths exhibit significant advantages over diagonal paths in dual-phalanx finger mechanisms\cite{Kim2022}. To achieve vertical compliance in our finger mechanism, we implement a Hoeckens linkage for linear motion generation.

\subsection{Fingertip Translation: Design of Offset Hoeckens Linkage for Vertical Compliance}

\begin {figure}[h]
    \centering
    \includegraphics [width=0.90\linewidth]{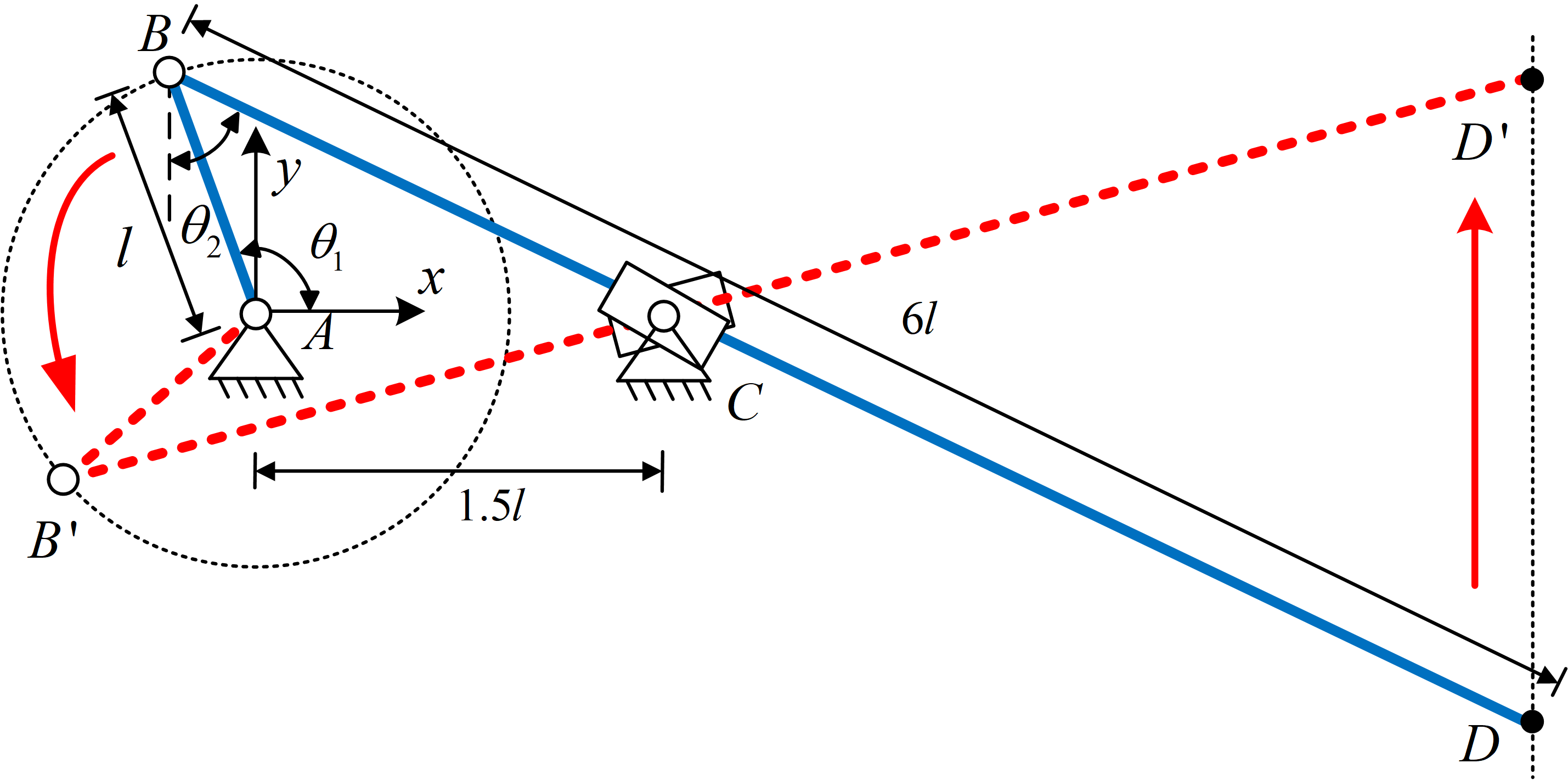}
    \caption {Motion Analysis of the Hoeckens Linkage}
    \label {fig:pathab}
\end {figure}

The kinematic model shown in Fig. \ref{fig:pathab} defines the geometric parameters where $l$ represents the unit length with $l_{AB}=l$, $l_{AC}=1.5l$, and $l_{BD}=6l$. Rotation angles $\theta_1$ and $\theta_2$ correspond to links $AB$ and $BD$ respectively. Additionally, Fig. \ref{fig:pathab} demonstrates the mechanism's linear motion during the rotation of $AB$ to $AB'$.

The fundamental closure equation in vector form is $\overrightarrow{l_{AB}} +\overrightarrow{l_{BC}} = \overrightarrow{l_{AC}}$. Expressed in Cartesian coordinates with the global frame fixed at point $A$ in Fig. \ref{fig:pathab}:

\begin{equation}
\begin{cases}
l_{AB} \cos \theta_1 + l_{BC} \sin \theta_2 = x_C  \\
l_{AB} \sin \theta_1 + l_{BC} \cos \theta_2 = y_C
\end{cases}
\label{eq:2}
\end{equation}

Treating $\theta_1$ as the independent variable, we derive $\theta_2$ and $l_{BC}$:

\begin{equation}
    \theta_{2}=\arctan\frac{y_C-l_{AB}\cos\theta_1}{x_C-l_{AB}\sin\theta_1}
    \label{eq:3}
\end{equation}

\begin{equation}
    l_{BC}=\sqrt{(x_A-l_{AB}\cos\theta_{1})^{2}+(x_B-l_{AB}\sin\theta_1)^2}
    \label{eq:4}
\end{equation}

The coordinates of critical point $D$ are determined by:

\begin{equation}
\begin{cases}
x_{D}= x_{C}+(l_{BD}-l_{BC})*\cos{\theta_2} \\
y_{D}= y_{C}+(l_{BD}-l_{BC})*\sin{\theta_2}
\end{cases}
\label{eq:5}
\end{equation}

Finally, we can consider that $\overrightarrow{AD} = f(\theta_1)$, which establishes a one-to-one correspondence between the coordinates of point D and $\theta_1$ through the kinematic chain.

\begin{figure}[h]
    \centering
    \includegraphics[width=0.99\linewidth]{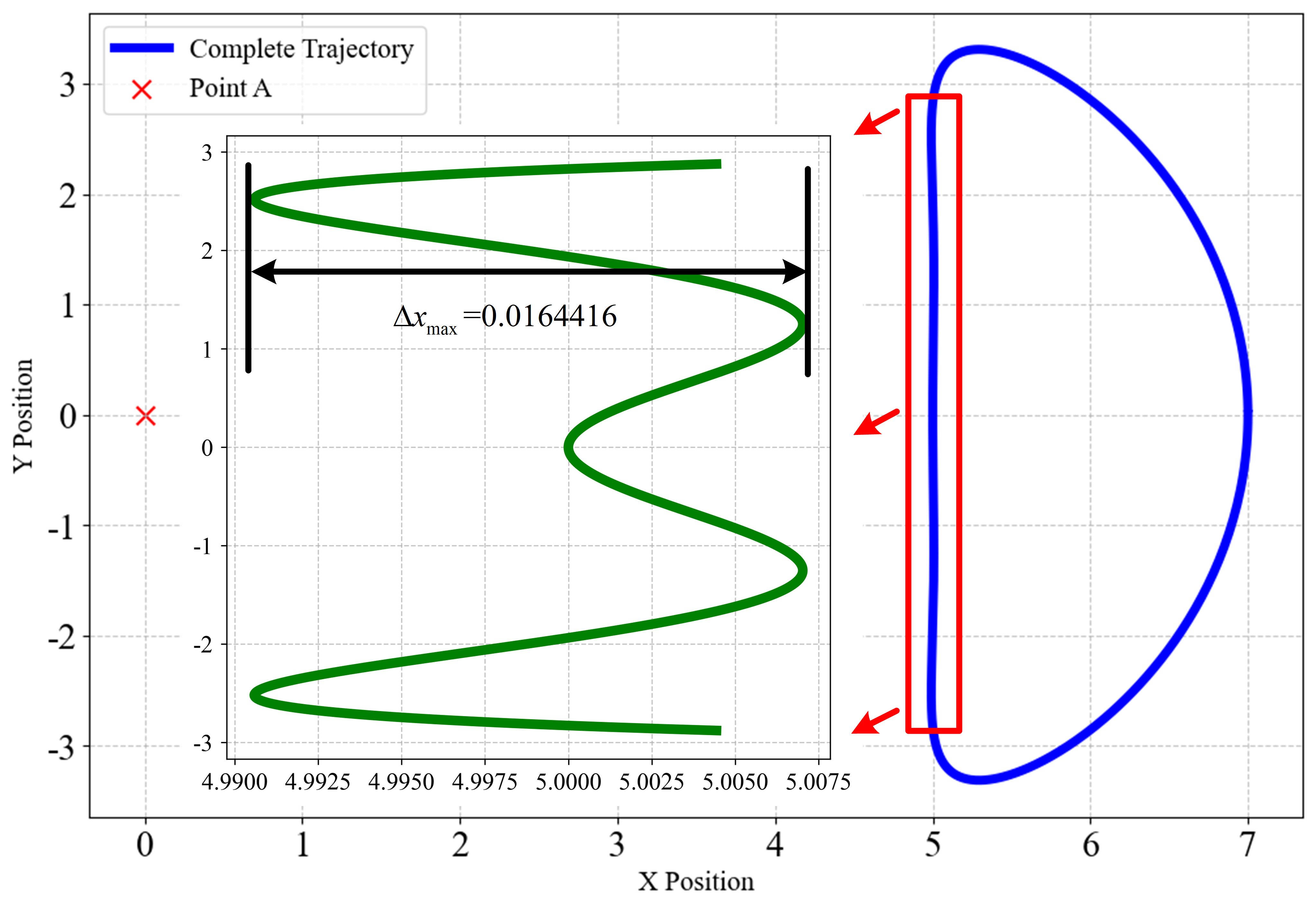}
    \caption{Displacement characteristics of point D during linkage operation}
    \label{fig:pathD}
\end{figure}

The displacement of point D was simulated, as shown in Fig. \ref{fig:pathD}. The mechanism achieves near-linear vertical motion between $\theta_1 = 68.51^\circ$ and $156.56^\circ$, with a maximum deviation of 0.0164 units, meeting precision requirements for robotic grasping.

\subsection{Fingertip Orientation: Design of Parallelogram Linkage and Four-bar Linkage}

\begin{figure}[h]
    \centering
    \includegraphics[width=0.9\linewidth]{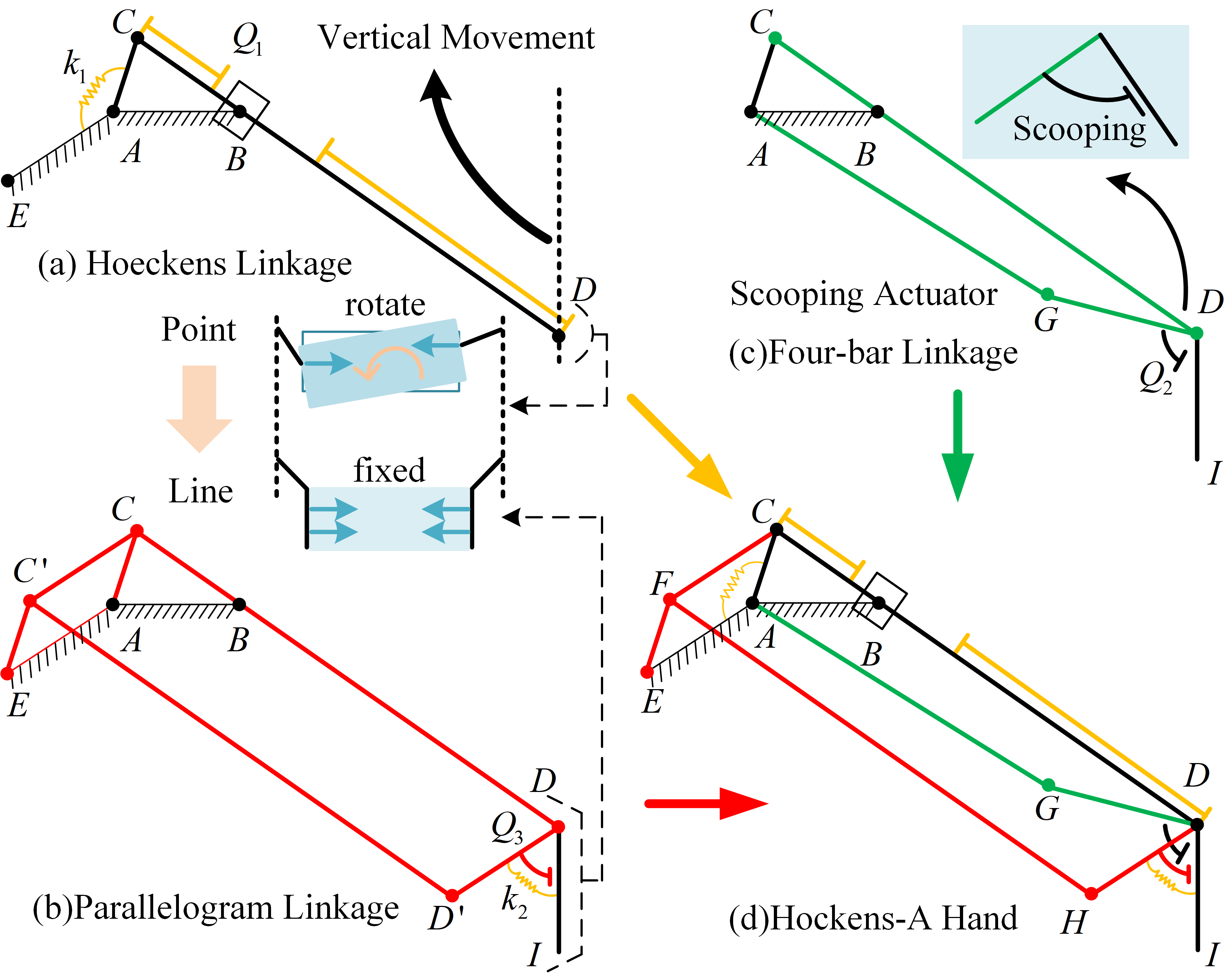}
    \caption{Mechanism evolution: (a) Basic Hoeckens linkage, (b) Parallelogram-augmented design, (c) Four-bar integrated system, (d) Final composite mechanism.}
    \label{fig:Combine}
\end{figure}

The parallelogram linkage forms a parallel contact interface with the object, offering significant advantages in stable pinch-grasping applications. However, the traditional Hoeckens linkage mechanism is limited in pinching operations due to its tip-point contact, which results in poor grasping stability and insufficient force distribution (see Fig. \ref{fig:Combine}(a)).

To address this limitation, we propose an improved double-parallelogram system(see Fig. \ref{fig:Combine}(b)). By adding an auxiliary base \(AE\) at \(150^{\circ}\) to \(AB\) with \(l_{AE}=l_{AB}\), and ensuring \(CC'\) is parallel to \(AE\), we construct two parallelogram linkages \(ACC'E\) and \(CC'D'D\). This system maintains the pure translational motion of the \(DD'\) rod, ensuring vertical translation with a constant orientation. The distal phalange \(DI\) is installed at the distal end of \(DD'\) to achieve stable line-contact pinching. The vertical stopper \(Q_3\) and return spring \(k_2\) ensure \(DI\) remains vertical in the non-driven state.

\begin{figure}[h]
    \centering
    \includegraphics[width=0.85\linewidth]{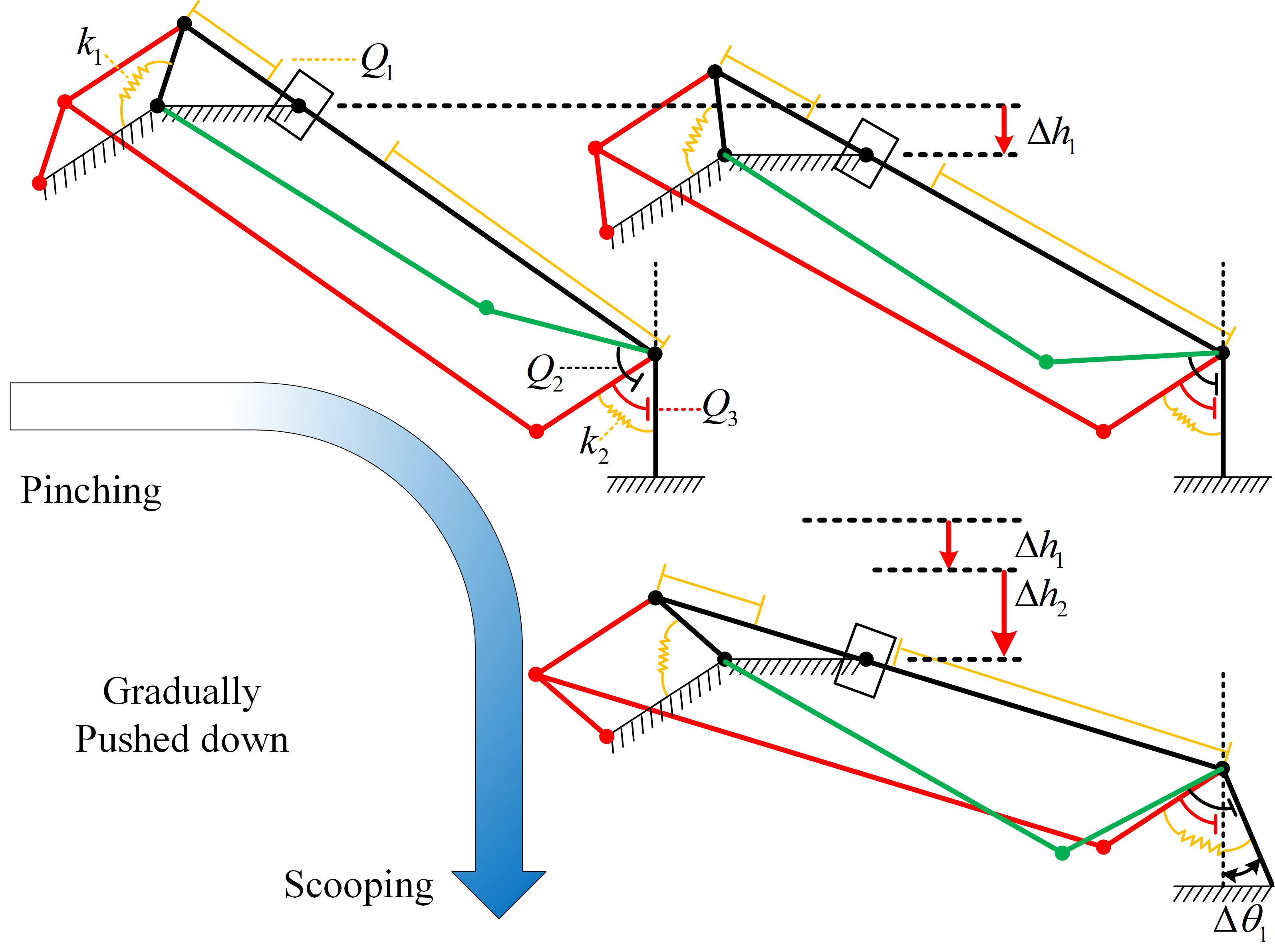}
    \caption{Posture transition sequence from pinching to scooping.}
    \label{fig:pushed}
\end{figure}

To enable passive switching between scooping and pinching postures, we integrate a four-bar trigger mechanism on \(DG\) and \(AG\) rods (see Fig. \ref{fig:Combine}(c)). This mechanism utilizes the idle-stroke gap of stopper \(Q_2\) for two-stage motion control. Initially, \(DI\) remains vertical. When the pressing displacement reaches \(\Delta h_1\) (\(161mm\leq h\leq180mm\)), \(Q_2\) contacts the trigger surface, initiating rotation around joint \(D\). As pressing continues, \(DI\) inclines outward by \(\Delta \theta_1 = 35^{\circ}\) at \(\Delta h_2\) (\(93mm\leq h < 161mm\)), transitioning from pinching to scooping (see Fig. \ref{fig:pushed}).

Fig. \ref{fig:Combine}(d) presents the final design of Hockens-A Hand,  while Table \ref{tab:combined-params} provides the detailed dimensional parameters of the design.

\begin{table}[ht]
    \centering
    \caption{Design Parameters: Rod Lengths, Angles, and Displacements}
    \label{tab:combined-params}
    \rowcolors{1}{blue!10}{yellow!20}
    \begin{tabular}{|c|c|c|c|c|c|}
        \hline
        Parameter & $l_{AC}$ & $l_{AG}$ & $l_{GD}$ & $l_{CD}$ & $l_{AB}$ \\ \hline
        Value & $30mm$ & $125mm$ & $50mm$ & $180mm$ & $45mm$ \\ \hline
        Parameter & $Q_2$ & $Q_3$ & $\Delta h_1$ & $\Delta h_2$ & $\Delta \theta_1$ \\ \hline
        Value & $81.5^{\circ}$ & $60^{\circ}$ & $19mm$ & $68mm$ & $35^{\circ}$ \\ \hline
    \end{tabular}
\end{table}

\subsection{3D Mechanism Embodiment: Compositional Modeling and Digital Prototype Realization}

\begin{figure}[h]
    \centering
    \includegraphics[width=0.82\linewidth]{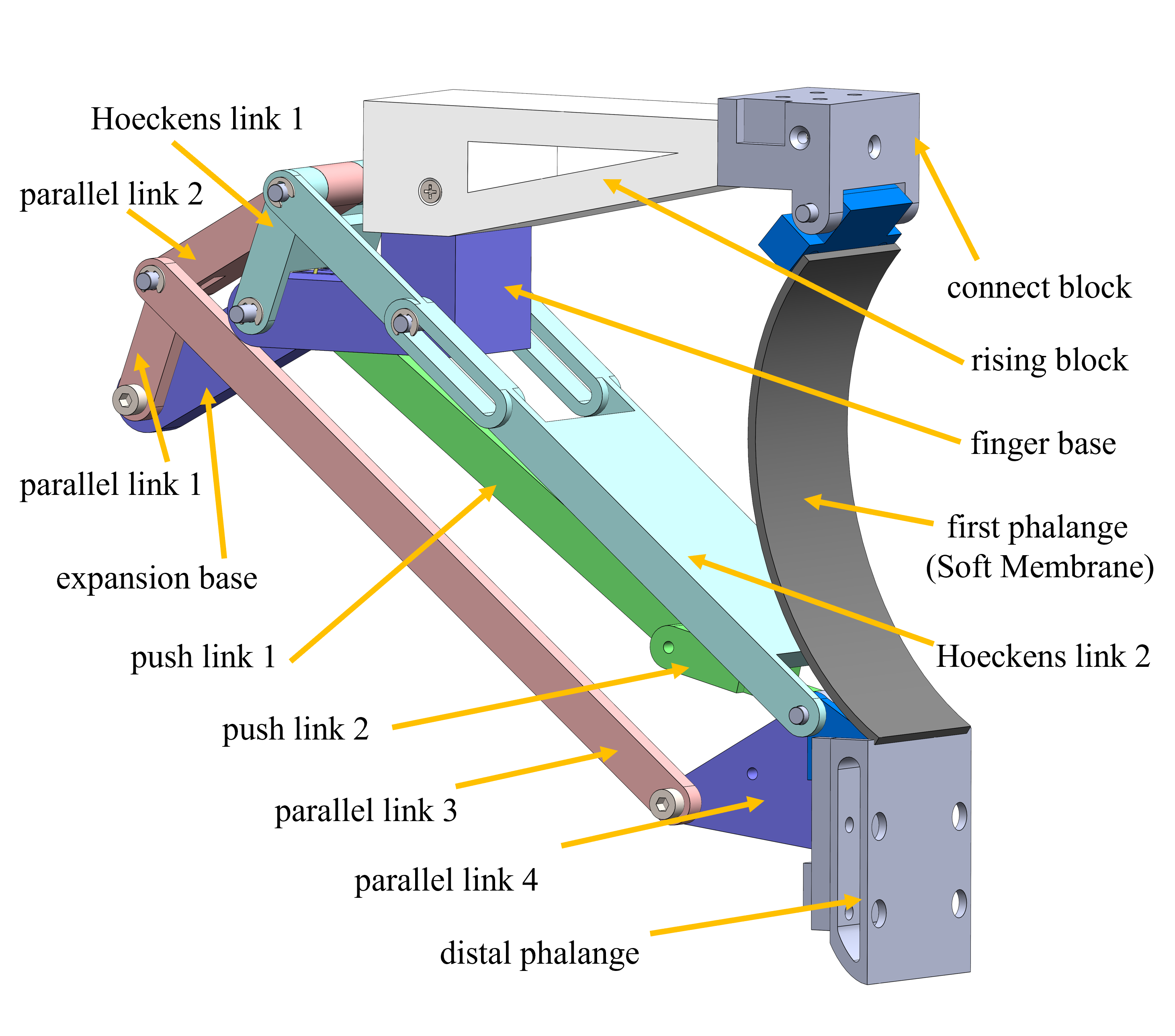}
    \caption{Configuration of the Hockens-A Hand.}
    \label{fig:whole_hand}
\end{figure}

Through digital modeling, we refined the 3D details of the Hockens-A hand, which consists of two symmetrical fingers. Fig. \ref{fig:whole_hand} illustrates the full view of one of the bionic fingers.

The first phalange features a $40\times110mm$ textured soft membrane, such as silicone or belt. Its elastic deformation enables adaptive grasping of large objects. In this paper, we utilize a silicone sheet for its excellent elastic properties. Alternatively, this phalange can also be constructed using elastic structural components or a series of rigid segments connected in tandem. This phalange can be adapted for active extension or retraction through motor actuation\cite{BLTgripper}. Moreover, its unidirectional constraint design ensures that such actuation does not compromise the desktop collision functionality, ensuring system reliability.

The Hoeckens linkage, comprising the input crank (Hoeckens link 1) and output rocker (Hoeckens link 2), achieves an approximate linear vertical trajectory of $\pm0.49mm$ with the end-chute constraint. 

The Double Parallelogram Mechanism (light red) maintains a dynamic parallel relationship between parallel link 4 and the base which ensures the distal phalange remains vertical in the default state and resets stably after movement. 

The four-bar system (green) amplifies the rotation angle of Hoeckens link 2 from $30.04^{\circ}$ to $59.88^{\circ}$ via push link 1-2, driving the distal phalange to perform the outward push.



\begin{figure}[h]
    \centering
    \includegraphics[width=0.98\linewidth]{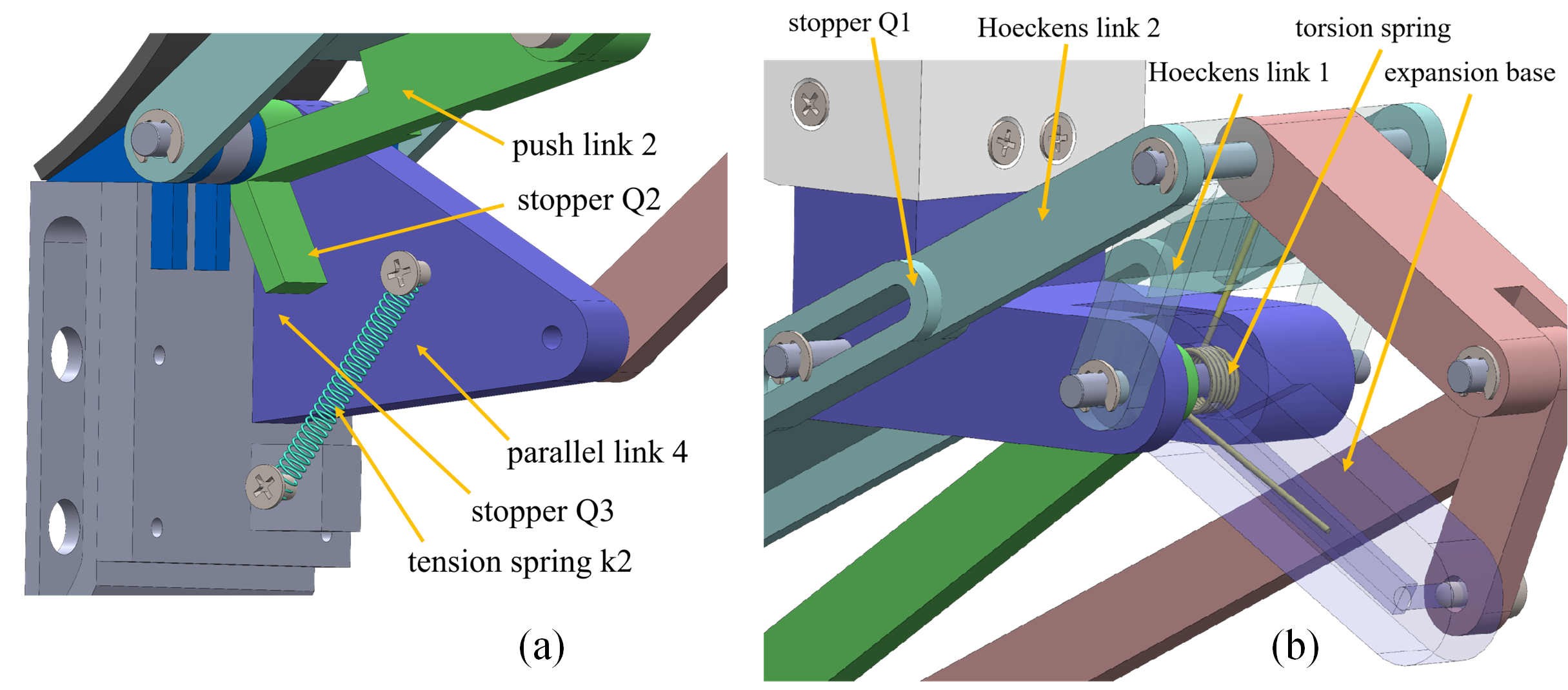}
    \caption{Mechanical details of the Hockens-A Hand: (a) Stopper \(Q_2\) and elastic reset mechanism of parallel link 4; (b) Torsion spring \(k_1\) installation on Hoeckens linkage.}
    \label{fig:k}
\end{figure}

As shown in Fig. \ref{fig:k}(a), the triangular block, which is also parallel link 4, integrates stopper \(Q_3\) and tension spring \(k_2\), ensuring the distal phalange adheres to \(Q_3\) under normal conditions. Push link 2 features stopper \(Q_2\), enabling motion stage separation via idle-stroke design.

In Fig. \ref{fig:k}(b), torsion spring \(k_1\) is installed on a 4mm steel shaft with a pre-tightening angle of $40^{\circ}$, maintaining the initial posture with the chute limit \(Q_1\).

\section{ANALYSIS}

\subsection{Kinematic Modeling and Parameter Optimization of Four-Bar Linkage}

Precise control of excavation angles forms the core technical requirement for adaptive excavation operations, where dynamic characteristics are significantly influenced by four-bar linkage parameters. Through parametric kinematic modeling, this study reveals the nonlinear mapping relationship between link dimensions and angular variation amplitudes, establishing theoretical foundations for mechanism optimization.

Based on the fixed pivot A(0, 0) (topological structure shown in Fig. \ref{fig:Combine}) and using the link lengths of AG and DG (\( L_{\text{AG}} \), \( L_{\text{DG}} \)) as design variables, the kinematic model is constructed. By geometric resolution of intersecting circles, we can obtain two circles centered at A and D respectively, as follows:

\begin{equation}
    \begin{cases}
    X^2 + Y^2 = L_{\text{AG}}^2 \\
    (X - D_x)^2 + (Y - D_y)^2 = L_{\text{DG}}^2 \\
    \end{cases}
\end{equation}

The analytical solution for intersection point G derives as:

\begin{equation}
    \begin{aligned}
        d = \sqrt{D_x^2 + D_y^2}, a &= \frac{L_{\text{AG}}^2-L_{\text{DG}}^2 + d^2}{2d},h = \sqrt{L_{\text{AG}}^2-a^2},\\
        G_{1,2} &= \left( \frac{aD_x \pm hD_y}{d},\ \frac{aD_y \mp hD_x}{d} \right),
    \end{aligned}
\end{equation}

where the solution with smaller y-coordinate is selected per mechanical assembly requirements.

\begin{figure}[h]
    \centering
    \includegraphics[width=0.95\linewidth]{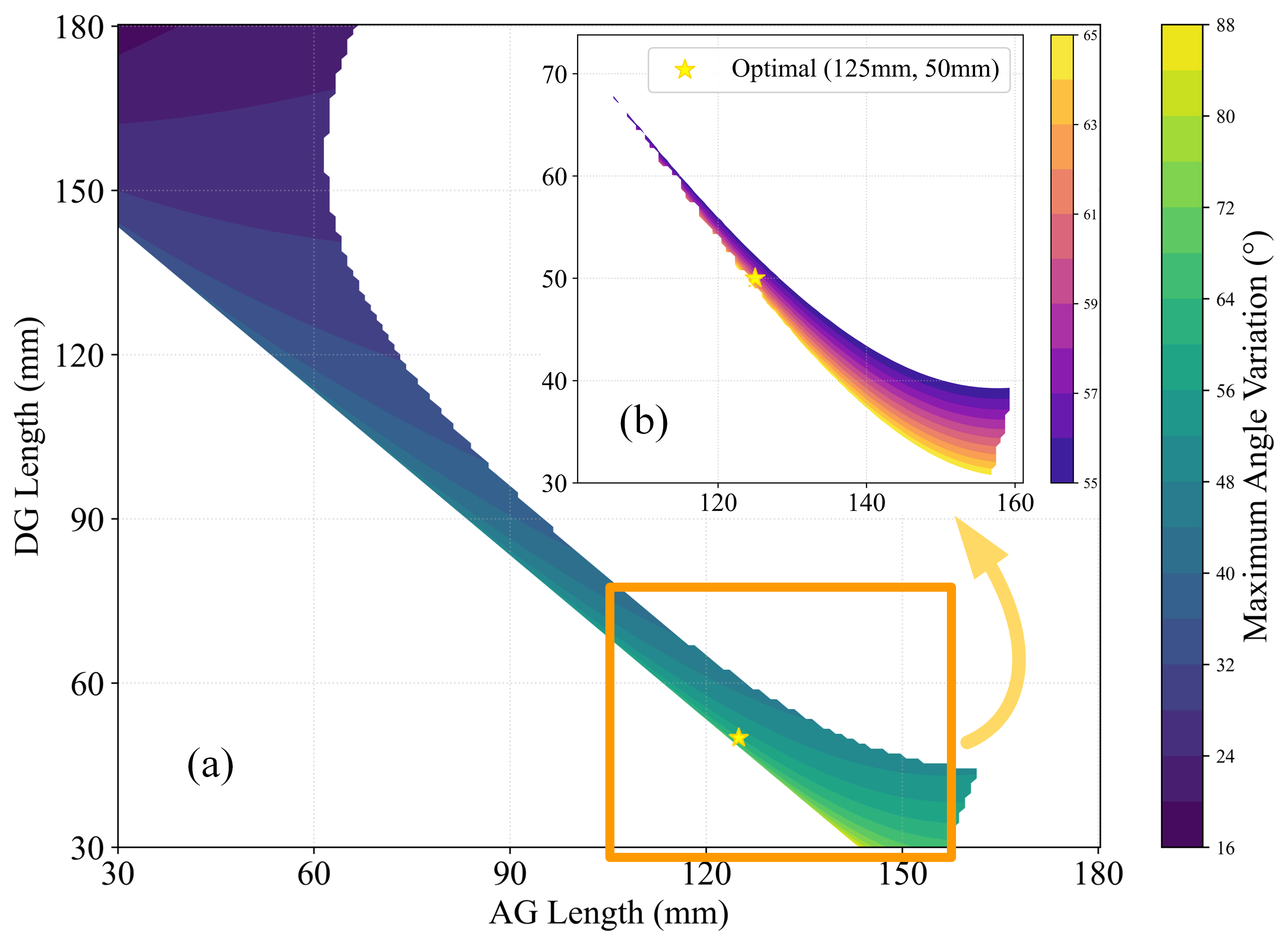}
    \caption{Four-bar parameter optimization contours: (a) Global angular distribution revealing parametric sensitivity and constraint boundaries; (b) Local design window highlighting solution.}
    \label{fig:fourbar}
\end{figure}

Two critical constraints govern the mechanism's validity. The first is “Motion continuity” ensured by the Grashof criterion: \(|L_{\text{AG}}-L_{\text{DG}}| \leq d \leq L_{\text{AG}} + L_{\text{DG}}\), and the second is “Workspace limitation” to prevent mechanical interference: \(G_y \geq D_y - 45\). Parameter combinations violating either constraint are deemed infeasible.

The push angle \( \theta \), defined as the orientation of the DG link relative to the positive y-axis, is calculated as

\begin{equation}
    \theta = \arctan2\left(D_x-G_x,D_y-G_y \right) \times \frac{180}{\pi}.
\end{equation}

The angular variation extremum, \( \Delta\theta_{\max} \), represents the maximum difference between the highest and lowest values of \( \theta \) observed during the mechanism's motion. Mechanisms with \( \Delta\theta_{\max} > 180^\circ \) often exhibit discontinuous motion due to link crossover or dead-center positions.

Full parametric scanning (\( L_{\text{AG}} \in [30.0,180.0] \)mm, \( L_{\text{DG}} \in [30.0,180.0] \)mm, 1mm resolution) generates angular variation contours (Fig. \ref{fig:fourbar}). The optimal combination \( L_{\text{AG}}=125 \)mm, \( L_{\text{DG}}=50 \)mm achieves \( \Delta\theta_{\max}=59.82^\circ \pm 0.5^\circ \) with less than 1\% experimental error.

\textbf{Parametric sensitivity analysis} reveals distinct gradient behaviors: AG length demonstrates strong positive correlation (\( r=0.915 \)) with 0.3174°/mm slope, where 30mm increases yield 9.5° \( \Delta\theta_{\max} \) growth, ideal for coarse adjustment; DG length below 60mm shows high negative sensitivity (-0.62°/mm slope), enabling 6.2° \( \Delta\theta_{\max} \) increase per 10mm reduction for fine control.

\subsection{Kinematic Analysis of Fingertip}

\begin{figure}[h]
    \centering
    \includegraphics[width=0.95\linewidth]{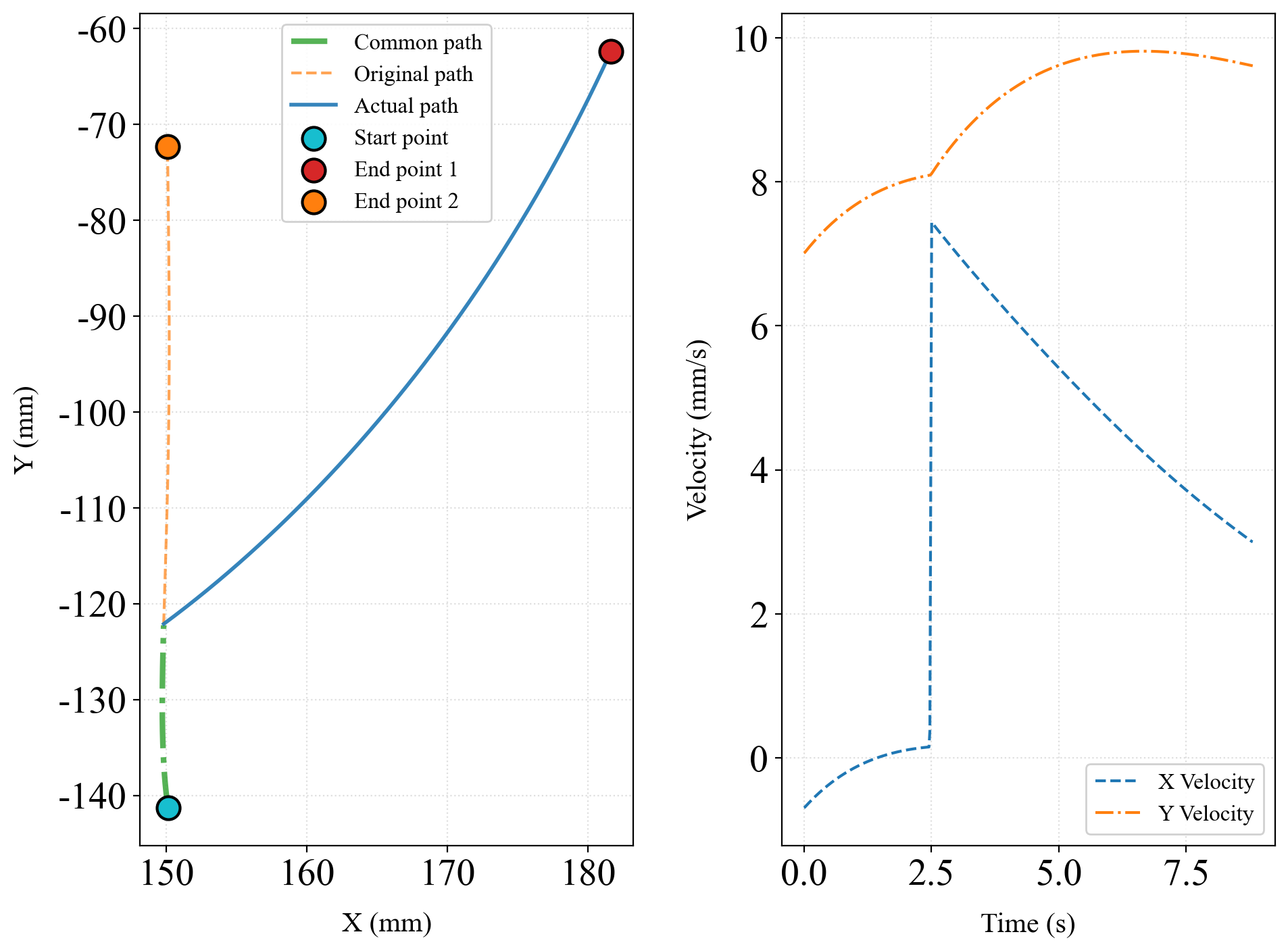}
    \caption{Trajectory of the fingertip: (a) XY coordinate changes showing the original and actual paths; (b) Velocity components in the X and Y directions.}
    \label{fig:G}
\end{figure}

Following the kinematic modeling and parameter optimization of the four-bar linkage, we further investigate the motion characteristics of the Hockens-A Hand. The simulation, conducted using Python, ultimately depicts the positional changes of the fingertip (see Fig. \ref{fig:G}).

From Fig.\ref{fig:G}(a), point I's XY trajectory exhibits two stages. Initially, both paths coincide for the first 16 mm due to the idle distance in the push design. Beyond this, the original path rises linearly to endpoint 1, while the pushed path, influenced by \(Q_2\), shifts horizontally to 182 mm and lifts vertically by 10 mm, reaching endpoint 2. In (b), the X-axis velocity shows a sudden increase to 7.438 mm/s at 2.506 s, marking the push-induced displacement in the X direction, followed by a gradual decline. The Y-axis velocity increases slightly during the push, then rises steadily until the end.

\begin{figure}[h]
    \centering
    \includegraphics[width=0.8\linewidth]{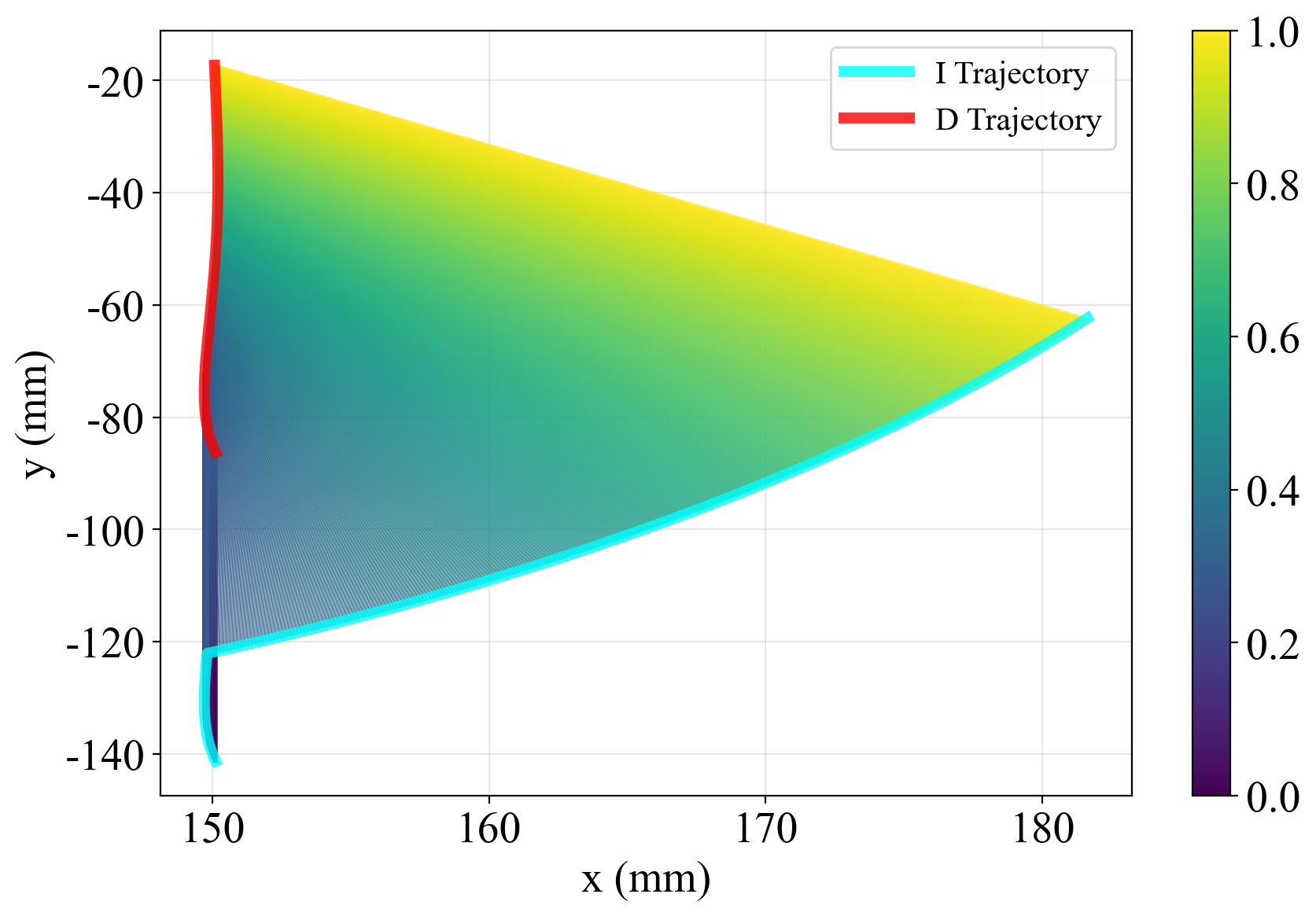}
    \caption{Complete motion trajectory of the distal phalange.}
    \label{fig:pace}
\end{figure}

In Fig.\ref{fig:pace}, the complete motion trajectory of the distal phalange is plotted, showing the process of the DI segment moving upward and then extending outward. To quantify the working range of the Hockens-A Hand, we use the Shoelace Formula to calculate the area covered by the path:

\begin{equation}
S_{area} = \frac{1}{2} \left| \sum_{i=1}^{n-1} (x_i y_{i+1}-x_{i+1} y_i) + (x_n y_1-x_1 y_n) \right|
\end{equation}

Through calculation, the area covered by the path is determined to be \textbf{153.95 mm²}. This result indicates that the 55 mm distal phalange can cover a relatively large working range with only vertical motion, fully demonstrating the compactness and efficiency of the Hockens-A Hand. 

\subsection{Grasping Force Analysis}

In the context of the established kinematic model, we further extend our analysis to the force dynamics during grasping operations. By synthesizing and applying the previously derived motion relationships, we employ the power equation method to investigate the variation of grasping force $F_N$ during fingertip grasping.

The instantaneous power of the system can be expressed as:
\begin{equation}
P_{press} = P_{k_1} + P_{k_2} + P_{DI}
\end{equation}
where
\begin{equation}
\begin{aligned}
P_{k_1} = \frac{dU_{k_1}}{dt} = k_1 \cdot \Delta\theta_{1} \cdot \omega_1 \\
P_{k_2} = \frac{dU_{k_2}}{dt} = k_2 \cdot \Delta x_1 \cdot v_x \\
P_{DI} = F_{N} \cdot v_{D} + F_{N} \cdot r \cdot \omega_2 \\
\end{aligned}
\end{equation}

Based on Equation \ref{eq:2}-\ref{eq:5} from Chapter 2, where $\overrightarrow{AD} = f(\theta_1)$, we derive the motion simulation of point D in the Hoeckens linkage, obtaining:\(v_D = \frac{df(\theta_1)}{dt} = \omega_1 f'(\theta_1)\)

From the kinematic modeling of the four-bar linkage and push angle in the previous section, we established the nonlinear relationship \(\theta_2 = g(\theta_1)\). Taking the time derivative of both yields: \(\omega_2 = \omega_1 g'(\theta_1)\).

The calculation of $x_1$ incorporates the four-bar linkage mechanism to determine the coordinates of the spring's end points, with \(x_1 = h(\theta_2)\), \(v_x = \frac{dx}{dt} = \omega_1 h'(\theta_1)\).

Regarding $P_{DI}$, $F_N \cdot v_D$ represents the power of the grasping force in the vertical direction, while $F_{N} \cdot r \cdot \omega_2$ denotes the rotational power of the grasping force, collectively forming the output power of the grasping mechanism.

Consequently, the instantaneous power equation can be simplified and reformulated as:
\begin{equation}
F_N = \frac{\frac{P_{press}}{\omega_1} - k_1 \Delta\theta_1 - k_2 \Delta x_1 \cdot \frac{dx_1}{d\theta_1}}{f'(\theta_1)  + r \cdot g'(\theta_1)}
\end{equation}

\begin{figure}[h]
    \centering
    \includegraphics[width=0.98\linewidth]{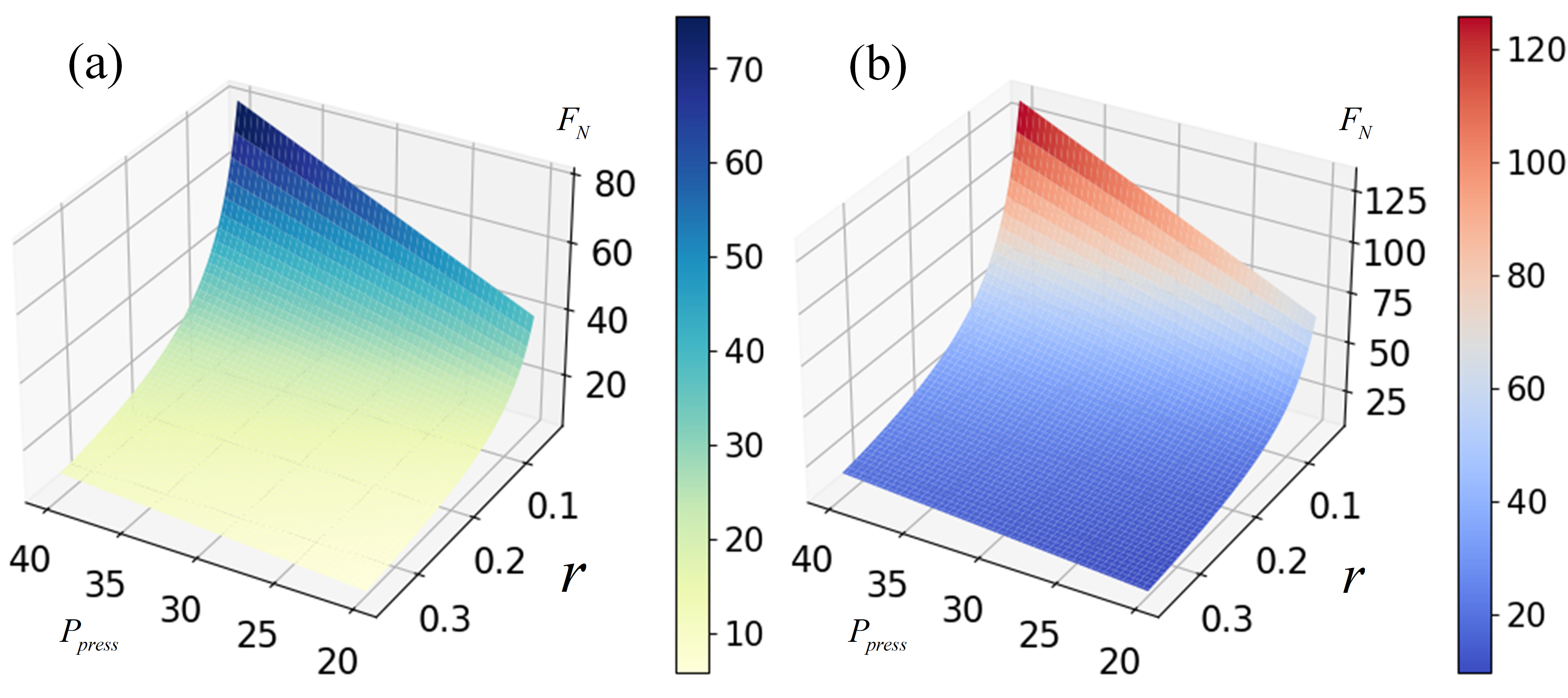}
    \caption {Analysis of \( F_N \) under variations of pressing power \( P_{\text{press}} \), distance \( r \) and rotation angle \(\theta_1\): (a) $\theta_1 = 80^\circ$; (b) $\theta_1 = 120^\circ$.}
    \label{fig:press}
\end{figure}

With the rotational velocity $\omega_1$ set to 10 degrees/s, the resulting Fig. \ref{fig:press}(a)(b) respectively illustrate the system configurations at $\theta_1 = 80^\circ$ and $\theta_1 = 120^\circ$. While the shapes of the two plots are nearly identical, the range of $F_N$ values is similar, but their starting points differ significantly. Specifically, the $F_N$ values in the $120^\circ$ plot are consistently larger than those in the $80^\circ$ plot. These results demonstrate that the $F_N$ is positively correlated with the $P_{press}$. Additionally, the pressure decreases as the position approaches the fingertip.

\section{EXPERIMENTS}

To rigorously and scientifically verify the feasibility of the design of Hockens-A Hand, we fabricated a 3D prototype. The robotic hand was disassembled into individual parts, printed using a Bamboo Lab A1 or P1S printer with PLA material. Actual measurements show that the robotic hand achieves a linear grasping range of 0-122 mm and a vertical lifting height of 87 mm.

\begin{figure}[h]
    \centering
    \includegraphics[width=0.98\linewidth]{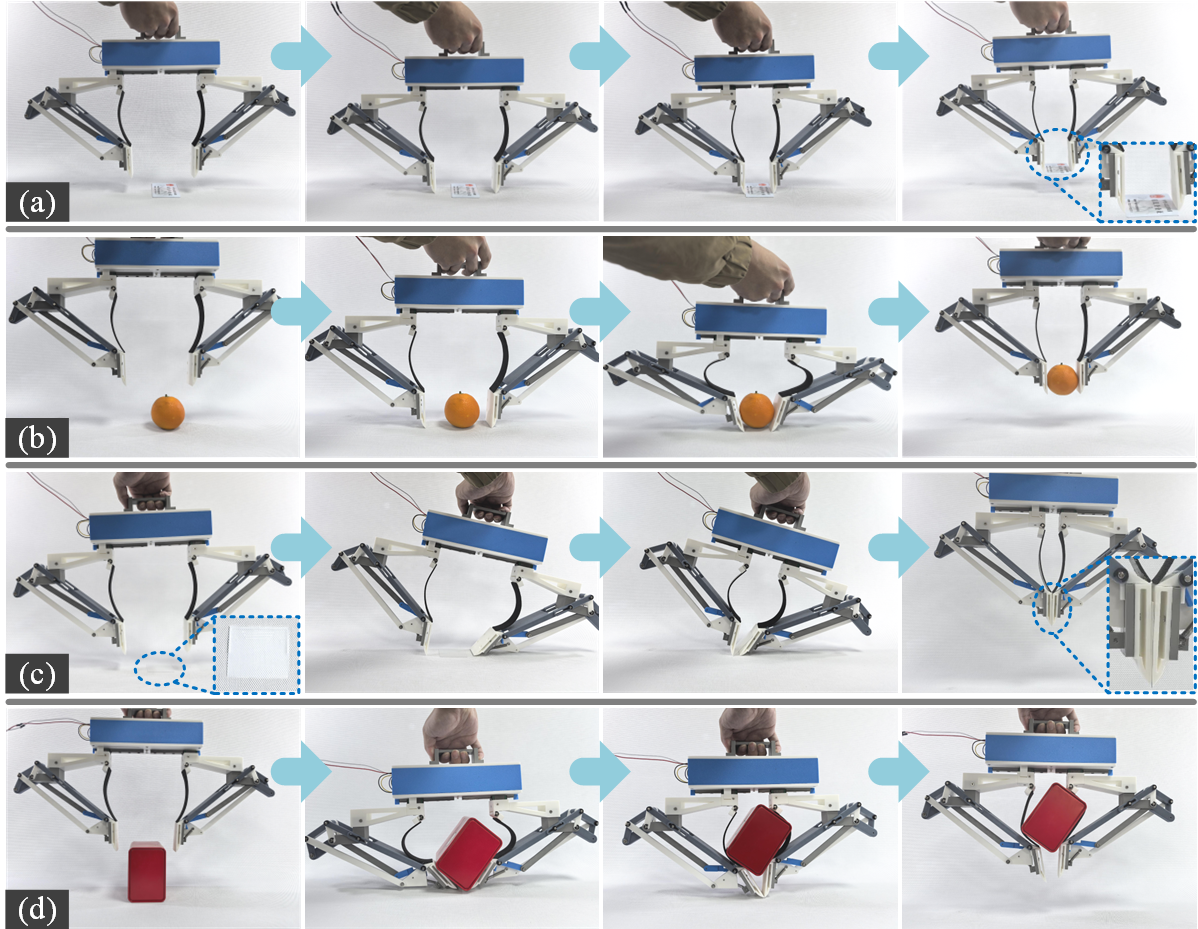}
    \caption {Experiments of Hoeckens-A Hand: (a) and (b) parallel-pinching of an ID card and an orange; (c) asymmetric scooping of a 0.5mm-thick polyethylene sheet; (d) symmetric scooping and stable enveloping of a 74x110x105-mm tea can.}
    \label{fig:grasp}
\end{figure}

In object-grasping tasks, for conventional objects, Hockens-A Hand designed in this study, similar to common robotic hands, can use the linear parallel-pinch method. Stable gripping is achieved through line-contact between the end-fingertips and the object.Fig. \ref{fig:grasp}(a)(b) show parallel-pinching of an ID card and an orange.

When handling thin sheets, the traditional parallel-pinch method suffers from instability, high failure rates, and random postures due to line-contact shifting to point-contact, causing rotation or slippage. This study introduces an asymmetric scooping strategy: one hand acts as a “Blocking” hand to push the object, while the other serves as a “Scooping” hand to lift and hold it. Fig. \ref{fig:grasp}(c) demonstrates stable grasping of a 0.5 mm polyethylene sheet.

When handling large irregular objects, parallel-pinching often results in slippage and instability due to limited gripping force and irregular shapes. To address this, the study introduces a method combining symmetric scooping and silicone enveloping. The object is lifted from the bottom and adaptively enveloped by flexible silicone, ensuring stable grasping. Fig. \ref{fig:grasp}(d) demonstrates this with a tea can.

Experimental results demonstrated that the asymmetric scooping strategy boosted the success rate for grasping thin sheets, like a 0.5 mm polyethylene sheet, to over 88\%, compared to the unstable traditional parallel-pinch method. Similarly, the symmetric scooping and silicone enveloping approach achieved a 90\% success rate for large irregular objects by adapting to their shapes and preventing slippage.

\begin{figure}[h]
    \centering
    \includegraphics[width=0.8\linewidth]{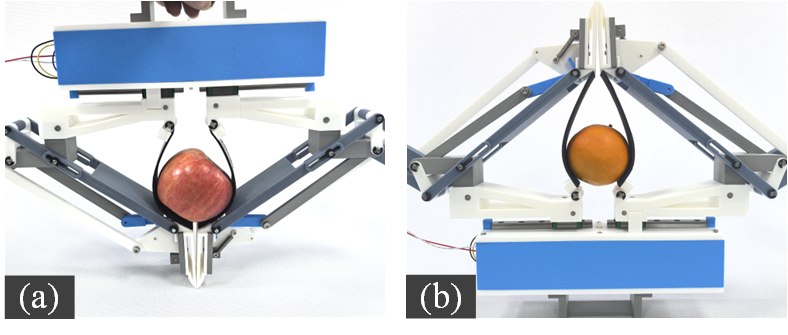}
    \caption{Experiments of the enveloping ability of soft silicone.}
    \label{fig:bao}
\end{figure}

The enveloping ability of the soft silicone was extensively tested across a range of objects with varying sizes and shapes. The results showed that the silicone could effectively adapt to and securely grip objects within a diameter range of 60 mm to 100 mm, achieving optimal enveloping performance. Outside this range, while the silicone still demonstrated some adaptability, the stability of the grasp was slightly reduced. As illustrated in Fig. \ref{fig:bao}, the successful enveloping of various objects visually demonstrates the silicone's effectiveness in improving grasping stability.







\section{CONCLUSIONS}
The Hockens-A Hand presented in this study introduces four key innovations in underactuated robotic grasping. First, its mechanical design employs passive intelligence to adapt to environmental constraints, ensuring stable grasping through intrinsic linkage interactions. Second, mesh-textured silicone phalanges enable adaptive enveloping of diverse objects, from thin sheets to irregular shapes, leveraging material flexibility. Third, the system achieves multifunctionality with a single linear actuator, reducing complexity and cost while maintaining adaptability in unstructured environments. Fourth, it combines three grasping modes: parallel pinching, symmetric scooping, and asymmetric scooping, overcoming limitations of traditional grippers. Experimental validation with a 3D-printed prototype demonstrated robust and precise grasping of objects ranging from 0.5-mm sheets to 105-mm cans. These advancements provide a compact, cost-effective solution for versatile manipulation in real-world applications.

\addtolength{\textheight}{-12cm}   




\end{document}